\documentclass[journal]{IEEEtran}

\usepackage{graphicx}
\usepackage{booktabs}
\usepackage{array}
\usepackage{graphics}
\usepackage{multirow}
\usepackage{utfsym}
\usepackage{graphicx}
\usepackage{epstopdf}
\usepackage{upgreek}
\usepackage{amsmath}
\usepackage{amsfonts}
\usepackage{subcaption}
\usepackage{tabularray}
\usepackage[numbers,sort&compress]{natbib}
\usepackage[colorlinks,linkcolor=black]{hyperref}
\usepackage{bm}
\usepackage{cancel}
\usepackage{algorithm}
\usepackage{algpseudocode}
\usepackage{amsthm}

\graphicspath{{pix/}}
 
\newcommand{\dt}[1]{\iffalse{#1}\fi}

\newcommand{\dtt}[1]{\iffalse{#1}\fi}

\usepackage{xcolor}

\usepackage{amssymb}
\usepackage{pifont}
\usepackage{booktabs}

\begin{document}

\title{Utilizing the Score of Data Distribution for Hyperspectral Anomaly Detection}

\author{Jiahui~Sheng, Yidan~Shi, Shu~Xiang, Xiaorun~Li and Shuhan~Chen
	
	\thanks{Preprint.}
}


\maketitle

\begin{abstract}
Hyperspectral images (HSIs) are a type of image that contains abundant spectral information. As a type of real-world data, the high-dimensional spectra in hyperspectral images are actually determined by only a few factors, such as chemical composition and illumination. Thus, spectra in hyperspectral images are highly likely to satisfy the manifold hypothesis.
Based on the hyperspectral manifold hypothesis, we propose a novel hyperspectral anomaly detection method (named ScoreAD) that leverages the time-dependent gradient field of the data distribution (i.e., the score), as learned by a score-based generative model (SGM).
Our method first trains the SGM on the entire set of spectra from the hyperspectral image. At test time, each spectrum is passed through a perturbation kernel, and the resulting perturbed spectrum is fed into the trained SGM to obtain the estimated score.
The manifold hypothesis of HSIs posits that background spectra reside on one or more low-dimensional manifolds. Conversely, anomalous spectra, owing to their unique spectral signatures, are considered outliers that do not conform to the background manifold. Based on this fundamental discrepancy in their manifold distributions, we leverage a generative SGM to achieve hyperspectral anomaly detection. 
Experiments on the four hyperspectral datasets demonstrate the effectiveness of the proposed method.
The code is available at https://github.com/jiahuisheng/ScoreAD.
\end{abstract}


\IEEEpeerreviewmaketitle

\section{Introduction}

\IEEEPARstart{D}{ifferent} from the visible image, hyperspectral images (HSIs) possess hundreds of channels, which can provide much more spectral information of the object than other formats of image~\cite{hyperspectral-background-1,hyperspectral-background-2}. 
Due to the high spectral resolution of hyperspectral images, there are many applications in different fields, such as hyperspectral unmixing~\cite{unmixing-background-1}, \cite{VCA}, hyperspectral classification~\cite{classification-background-1}, \cite{classification-background-TCSVT-2}, \cite{classification-background-TCSVT}, band selection \cite{BS-background-TCSVT-1}, \cite{BS-background-TCSVT-2} and hyperspectral anomaly detection~\cite{HAD-background-1, HAD-background-2}. 
Among these applications, hyperspectral anomaly detection is undoubtedly one of the hotspots and has attracted significant attention. 

Hyperspectral anomaly detection (HAD), which can be viewed as a pixel-wise binary classification task without prior spectral information~\cite{HAD-background-1}. We need to detect potentially anomalous spectra in hyperspectral images without labeled data, which is challenging. For anomalies in hyperspectral images, we generally consider them to have two primary characteristics: 1. They possess spectra that are distinctly different in statistical distribution from other pixels; 2. They occupy a very small proportion in the hyperspectral image. 

Over the past two decades, numerous researchers have devoted significant effort to this field, resulting in the development of many effective detection methods. Generally, these methods can be broadly categorized into four types: 
\begin{enumerate}[]
	
	\item Statistics-based methods, primarily represented by the Reed-Xiaoli anomaly detector (RXAD) \cite{RXAD} and its variants, such as LRX \cite{LRX}, KRX \cite{KRX}, and FrRX \cite{FrRX}.
	
	\item Representation-based methods, including approaches based on low-rank sparse representation \cite{LRASR}, \cite{LRASR_BP}, \cite{LRASR_UD}, \cite{SD_LRASR} and collaborative representation \cite{CRD}, \cite{CR_PCA}, \cite{RCRD}.

	\item Decomposition-based methods, divided into matrix decomposition-based \cite{LSMAD}, \cite{LRaSMD}, \cite{MX-SVD}, \cite{LSDM-MoG} and tensor decomposition-based approaches \cite{Tucker}, \cite{TensorRPCA}, \cite{PBTA}.

	\item Deep learning-based methods, encompassing techniques based on CNNs \cite{CNN-based}, GANs \cite{GAN-based}, Transformer \cite{TAEF}, \cite{GT-HAD}, and Autoencoders (AEs) \cite{Auto-AD}, \cite{LREN}, \cite{HAD-Dual-Net}, \cite{Memory-Augmented-AE}.
	
\end{enumerate}

Besides the four primary categories mentioned above, there are other types of HAD methods, like graph-based methods \cite{graph-based-1}, \cite{graph-based-2}, \cite{Graph_Frequency-HAD}, \cite{IGPAD}, topology-based methods \cite{topology-based}, \cite{chessboad-topology}, and manifold-based methods\cite{Embedding_Manifold_AE_HAD}, \cite{Manifold_Ranking-Based_HAD}.

Here, manifold-based methods refer to those that utilize the manifold hypothesis and related techniques. Given that hyperspectral images adhere well to this hypothesis, manifold has been extensively applied to tasks like classification \cite{manifold-HSIC-1}, \cite{manifold-HSIC-3}, unmixing \cite{manifold-unmixing-1}, and band selection \cite{manifold-BS-1}. Nevertheless, to our knowledge, the application of these methods has been largely superficial. They predominantly incorporate manifold-related techniques (e.g., manifold ranking \cite{manifold-ranking-HSIC-1}, manifold clustering \cite{manifold-unmixing-3}, regularization term \cite{manifold-grassmann-clustering}, \cite{manifold-constrain-PCA}) as auxiliary components to boost model performance, rather than designing applications based on the intrinsic properties of the manifold. This tendency is particularly evident in the domain of hyperspectral anomaly detection \cite{manifold-HAD-1}.

This situation can be largely attributed to a primary challenge: the intractability of obtaining an explicit representation of the hyperspectral data manifold. This has greatly limited the further application of manifolds in the hyperspectral field, especially in the domain of hyperspectral anomaly detection. Therefore, we identify two fundamental challenges that hinder the development of a truly manifold-based anomaly detection method:

\begin{enumerate}[]
	
	\item  How to obtain the representation of the manifold of hyperspectral image.
	
	\item Once represented, how to effectively utilize the manifold structure to discriminate between background and anomalous spectra.
	
\end{enumerate}

Regarding the first challenge, obtaining an explicit mathematical representation of the data manifold is often intractable due to its inherent complexity in high-dimensional spaces. An alternative to explicitly approximating the manifold is to learn it implicitly through a generative model \cite{stanczuk2024diffusion}. A generative model that successfully synthesizes data points consistent with the target distribution is considered to have effectively learned the underlying data manifold \cite{DDPM}.
The second challenge involves differentiating between background and anomaly. We posit that background spectra are points residing on the data manifold, whereas anomalies are points distant from it. This distributional differences can be captured by the score function—the gradient of the data's log-probability density. For an on-manifold point (background spectrum), the score field in its neighborhood will point towards the manifold from multiple directions. Conversely, for an off-manifold point (anomalous spectrum), the local score field will generally point toward the manifold's location from a single, dominant direction. This discrepancy in the orientation of local score field provides a robust criterion for discrimination. Consequently, the task requires a generative model capable of estimating the score function. Fortunately, the score-based generative model perfectly fits our requirements \cite{song2020score}.

Based on the score-based generative model, we propose a novel manifold-based hyperspectral anomaly detection method (named ScoreAD). 
The specific steps of ScoreAD are as follows. First, we trained a score-based generative model (SGM) using the entire set of spectra from the hyperspectral image, which subsequently acts as our score estimator. In the subsequent step, each spectrum from the HSI is perturbed via a Gaussian kernel to generate a corresponding data point, whose score is then computed by the SGM. This process is repeated multiple times for each original spectrum. By generating a sufficient number of these perturbed points and their associated scores, we can effectively estimate the score distribution in the local neighborhood of each spectrum. For background spectra (normal spectra), which lie on the manifold, the directions of the scores around them will be quite diverse, with various orientations present. For anomalous spectra, since they are distant from the data manifold, the directions of the scores around them will be highly concentrated and consistent. Based on this difference, we can achieve hyperspectral anomaly detection.

The main contributions of this paper can be summarized as follows:

\begin{enumerate}[]

  \item We innovatively treat background spectra as on-manifold points and anomalous spectra as off-manifold points. Based on the manifold structure itself, this allows us to achieve the detection of anomalous targets in hyperspectral images.
  \item We demonstrate that for points on the manifold and points off the manifold, the score fields within their respective neighborhoods have different flow directions. Based on the direction of the score field in a point's neighborhood, we can distinguish whether the point is on or off the manifold.
  \item We leverage a score-based generative model to serve a dual purpose: first, to implicitly learn the manifold of the hyperspectral background, and second, to estimate the local score field for each test point.
	
\end{enumerate}

The organizations of the paper are as follows. 
Section II reviews the applications of manifolds in hyperspectral image processing.
Section III provides a brief introduction to manifold hypothesis of hyperspectral image and score-based generative model. 
In Section IV, we present a detailed description of our proposed ScoreAD. 
In Section V, we conducted comparative experiments for ScoreAD.
Finally, we concludes the paper in Section VI.

\section{Related Work}
Manifolds are widely used in hyperspectral image processing. Next, we will introduce the common manifold-based techniques used in the hyperspectral domain.

\subsection{Manifold-based hyperspectral Image Processing}
Due to the strong suitability of manifolds for hyperspectral imagery, they have been widely used in various hyperspectral imaging applications, including classification, unmixing, band selection, and anomaly detection. Typically, manifolds are used to ensure the local smoothness of the hyperspectral data. For instance, \cite{manifold-HSIC-1} utilized a supervised local manifold learning algorithm to compute the similarity weights between samples, thereby achieving hyperspectral classification without dimensionality reduction. In \cite{manifold-HSIC-2}, the authors constrained a graph convolutional network by introducing a manifold geometry regularization term, enhancing the hyperspectral classification performance. 
Furthermore, various manifold-based techniques, including manifold ranking \cite{manifold-ranking-HSIC-1}, manifold clustering \cite{manifold-unmixing-3}, manifold preserving constraints \cite{manifold-BS-1}, and graph Laplacian regularization terms \cite{manifold-unmixing-1}, have been extended to hyperspectral applications such as unmixing \cite{manifold-unmixing-1}, \cite{manifold-unmixing-2}, band selection \cite{manifold-BS-1}, \cite{manifold-BS-2}, and anomaly detection \cite{RGAE}, \cite{manifold-HAD-1}.

While manifold-based techniques are now prevalent in hyperspectral image processing, their application tends to be highly homogeneous and modular. Too often, they are superficially employed as a generic module—much like a joint spatial-spectral feature extractor—to be simply integrated into existing frameworks, rather than forming the basis of novel methods that fundamentally explore the hyperspectral manifold itself.

\section{Preliminaries}
In this section, preliminary knowledge of the proposed ScoreAD method will be introduced, including the hyperspectral manifold hypothesis and the score-based generative model. 

\subsection{The Manifold Hypothesis of Hyperspectral Image}
In this subsection, we briefly review the manifold hypothesis of hyperspectral image, which is the foundation of the proposed ScoreAD. You may skip this section if you are already familiar with the manifold hypothesis of hyperspectral image.

\textbf{Problem Definition.} The \textit{Manifold Hypothesis} posits that high-dimensional real-world data, such as natural images, is not uniformly distributed throughout the ambient space. Instead, these data points tend to concentrate on or near underlying low-dimensional structures known as manifolds.

For Hyperspectral images (HSIs), we extend this to a \textit{Multi-Manifold Hypothesis}. 
We posit that the spectra within an entire hyperspectral scene reside on a low-dimensional structure that is a composite of multiple sub-manifolds. 
Specifically, we argue that:
\begin{enumerate}
    \item The spectral signature of a single endmember class is governed by a limited number of physical and environmental factors, such as illumination and humidity \cite{unmixing-background-1}. Consequently, the set of all spectra corresponding to this class constitutes a low-dimensional sub-manifold.
    \item The global structure of the spectra within an entire HSI is formed by these individual sub-manifolds and the geometric structures that connect them, which arise from the physical process of spectral mixing between different endmember classes.
\end{enumerate}
The following formal argument substantiates this hypothesis.

\textbf{Definition 1.} An endmember, $\mathbf{e}_k \in \mathbb{R}^C$, is the pure spectral signature of a single, macroscopically distinct material (land cover). Spectral variability, refers to the phenomenon that the spectral signature of a single endmember class varies continuously due to several factors like illumination and atmospheric conditions.

\textbf{Assumption 1.} Our theoretical framework is built upon the assumption that the spectra in hyperspectral images can be effectively described by current physical models, like the spectral variability model for single material and the spectral mixing model for mixed pixels.

\textbf{Lemma 1.} \textit{The set of all ideal (noise-free) spectral signatures for a single land cover class $k$ forms a $d_k$-dimensional manifold (denoted $\mathcal{M}_k$) embedded in $\mathbb{R}^C$, where the intrinsic dimension $d_k$ is significantly smaller than the ambient dimension $C$.}

\begin{proof}
Let the spectral variability of land cover class $k$ be governed by $d_k$ continuous physical parameters $\mathbf{\theta}_k \in \Theta_k \subset \mathbb{R}^{d_k}$, where $d_k \ll C$. The observed spectrum is given by a smooth, continuous mapping $f_k: \Theta_k \to \mathbb{R}^C$ such that $\mathbf{x} = f_k(\mathbf{\theta}_k)$. The image of a smooth map from a low-dimensional space to a high-dimensional space is, by definition, a $d_k$-dimensional manifold. Thus, the set $\mathcal{M}_k = \{ f_k(\mathbf{\theta}_k) \mid \forall \mathbf{\theta}_k \in \Theta_k \}$ is a $d_k$-dimensional manifold.
\end{proof}

\textbf{Theorem 1.} \textit{The set of all ideal (noise-free) spectra in a HSI scene, composed of $p$ distinct land cover classes, reside on a composite, piecewise-smooth manifold $\mathcal{M}_{\text{global}}$. The intrinsic dimension of this global manifold, $d_{\text{global}}$, is significantly smaller than the ambient dimension $C$.}

\begin{proof}
From Lemma 1, we establish that each pure material class $k$ constitutes a low-dimensional sub-manifold $\mathcal{M}_k$. A mixed pixel is a combination of endmembers $\{\mathbf{e}_k\}_{k=1}^p$, which are drawn from these sub-manifolds. Under spectral mixing models (e.g., the Linear Mixing Model), the set of all mixed pixels forms a geometric structure (e.g., simplex) that connects the sub-manifolds $\{\mathcal{M}_k\}$. The union of the individual sub-manifolds and these connecting structures forms a composite, piecewise-smooth manifold. Its intrinsic dimension $d_{\text{global}}$ is determined by the dimensions of the sub-manifolds and the degrees of freedom in their mixing ($p-1$), and thus it holds that $d_{\text{global}} \ll C$.
\end{proof}

According to Theorem 1, we have konwn that in the theoretical, noise-free case, the spectra within a HSI scene form a composite, piecewise-smooth manifold. In practice, however, the presence of noise means that real-world spectra do not lie strictly on this manifold but are instead highly concentrated around it.

\begin{figure}[!h]
\centering
\includegraphics[width=0.9\linewidth]{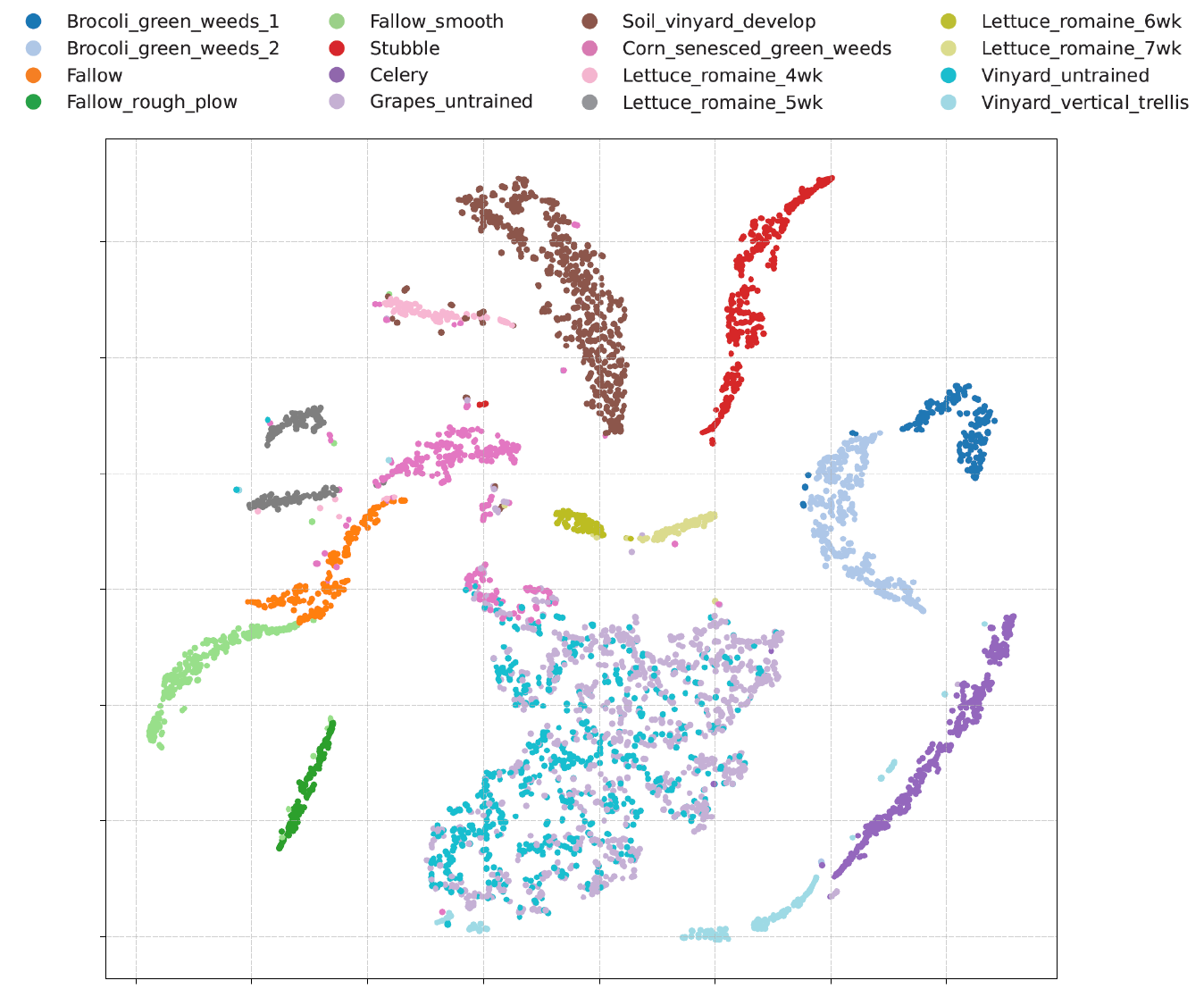}
\caption{
    t-SNE visualization of Salinas dataset.
}
\label{fig:manifold_visualization}
\end{figure}

\textbf{Visual Verification of the Manifold Hypothesis.} To provide empirical evidence for the manifold hypothesis, we visualize the spectra of Salinas dataset in Fig.~\ref{fig:manifold_visualization}. The figure shows that the spectra forms a composite low-dimensional manifold structure, supporting our theoretical claims. The visualization also illustrates how the spectral signatures of different land cover classes are clustered together, forming individual sub-manifolds.

\subsection{Score-based Generative Model}
Score-based generative model (SGM) is one type of the diffusion model that can generate high-fidelity samples by first gradually corrupting data into noise and then learning to reverse this process. The formulation of the SGM can be elegantly described under the framework of Stochastic Differential Equations (SDEs) \cite{song2020score}.

\subsubsection{The Score Function.}

The fundamental concept is the \textbf{score function} of a probability density $p(\mathbf{x})$, defined as the gradient of its log-density with respect to the data variable $\mathbf{x}$:
\begin{equation}
    \mathbf{s} (\mathbf{x}) = \nabla_{\mathbf{x}} \log p(\mathbf{x})
\end{equation}
The score function points in the direction of the steepest ascent of the log-density, effectively indicating how to modify a data point $\mathbf{x}$ to increase its likelihood. A neural network trained to estimate this function for a data distribution is called a score model.

\subsubsection{Forward Process: Corrupting Data with an SDE.}

The forward process systematically transforms a complex data distribution $p_0(\mathbf{x})$ into a simple, tractable prior distribution (e.g., a standard Gaussian distribution) over a continuous time interval $t \in [0, T]$. This can be achieved using a stochastic differential equation of the form:
\begin{equation}
    d\mathbf{x} = \mathbf{f}(\mathbf{x}, t) dt + g(t) d\mathbf{w}
    \label{eq:forward_sde}
\end{equation}
where $\mathbf{f}(\mathbf{x}, t)$ is a vector-valued function called the \textit{drift coefficient}, $g(t)$ is a scalar function called the \textit{diffusion coefficient}, and $\mathbf{w}$ is a standard Wiener process (Brownian motion). This SDE defines a continuous diffusion process that gradually adds noise to the data. Let the distribution of the process at time $t$, $\mathbf{x}_t$, be denoted by $p_t(\mathbf{x})$.

\subsubsection{Reverse Process: Generating Data with a Reverse-Time SDE.}

Remarkably, the reverse of this diffusion process can also be described by an SDE, known as the reverse-time SDE. This reverse SDE evolves from $t=T$ back to $t=0$ and transforms samples from the simple prior distribution $p_T$ back into samples from the original data distribution $p_0$. The reverse-time SDE is given by:
\begin{equation}
    d\mathbf{x} = [\mathbf{f}(\mathbf{x}, t) - g(t)^2 \nabla_{\mathbf{x}} \log p_t(\mathbf{x})] dt + g(t) d\bar{\mathbf{w}}
    \label{eq:reverse_sde}
\end{equation}
where $d\bar{\mathbf{w}}$ is a standard Wiener process running backward in time, and $dt$ is an infinitesimal negative time step.

Crucially, solving this reverse SDE to generate data requires knowing the score function, $\nabla_{\mathbf{x}} \log p_t(\mathbf{x})$, of the noisy data distribution at every intermediate time $t$.

\subsubsection{Training: Learning the Score Function.}

Since the true score function $\nabla_{\mathbf{x}} \log p_t(\mathbf{x})$ is unknown, it is approximated with a time-dependent, conditional neural network, $\mathbf{s}_\theta(\mathbf{x}, t)$, called the score model, with parameters $\theta$. This model is trained to estimate the true score across all time steps $t \in [0, T]$.

The parameters $\theta$ are optimized by minimizing a weighted sum of denoising score matching objectives over all time steps and data samples. The objective function is:
\begin{equation}
    \min_{\theta} \mathbb{E}_{\substack{{t\sim U(0,T)} \\ {\mathbf{x}_0\sim p_0(\mathbf{x}_0)} \\ {\mathbf{x}_t\sim p_t(\mathbf{x}_t|\mathbf{x}_0)}}}
[\lambda(t)\left\|\nabla_{\mathbf{x}_t}\ln p_t(\mathbf{x}_t|\mathbf{x}_0)-s_\theta(\mathbf{x}_t,t)\right\|_2^2]
    \label{eq:loss_function}
\end{equation}
Here, $\lambda(t)$ is a positive weighting function, and $\nabla_{\mathbf{x}_t} \log p_{t}(\mathbf{x}_t|\mathbf{x}_0)$ is the score of the perturbation kernel, which is tractable and can be easily computed as $-(\mathbf{x}_t - \mathbf{x}_0) / \sigma_t^2$ for Gaussian perturbations. Once the score model $\mathbf{s}_\theta$ is trained, it can be plugged into the reverse-time SDE (Eq. \ref{eq:reverse_sde}). By numerically solving this SDE starting from a random sample $\mathbf{x}_T \sim p_T$, we can generate new, high-fidelity data samples. 
In this work, however, the generation of new samples is unnecessary. The trained score model $\mathbf{s}_\theta$ is used directly and exclusively for predicting the scores of perturbed data.

Readers can refer to \cite{song2020score}, \cite{song2019generative} for further information.

\section{Proposed Method}

In this section, we will first introduce the principle of ScoreAD. Then, we will give a specific implementation of the ScoreAD on HAD.

\subsection{How the Score Differentiate Anomalous from Normal Data?}
\label{sec:theoretical_principle}

\begin{figure}[!h]
\centering
\includegraphics[width=\linewidth]{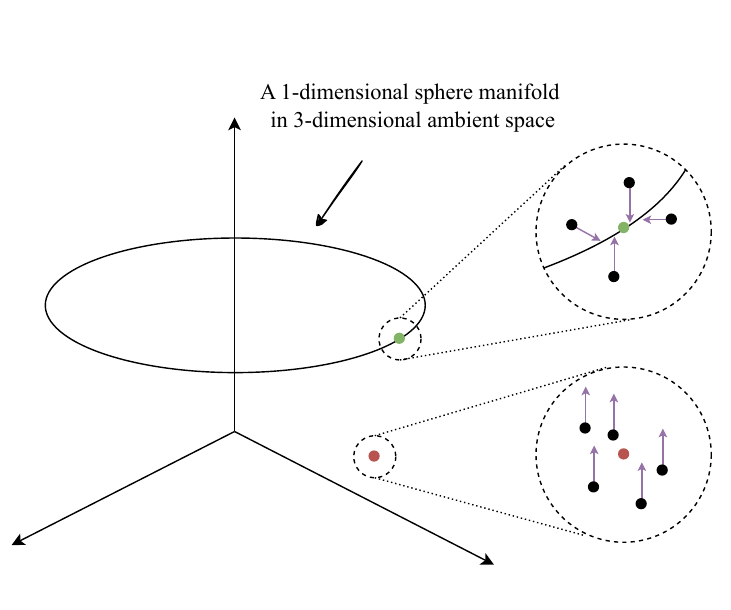}
\caption{
    Illustration of the differences between the score field around datapoint on the manifold and the score field around datapoint outside the manifold. The scores obtained from the perturbed datapoint of green datapoint (on manifold) are orthogonal to manifold and have diverse directions, while the scores around red datapoint have a consistent direction.
}
\label{fig:manifold_score}
\end{figure}

Our method is primarily based on the manifold hypothesis for high-dimensional data. This principle broadly applies to various types of natural data, such as natural visible image and spectra in hyperspectral image, which means our method theoretically possesses a degree of generality. Therefore, in this section, we will introduce our method in a more general context.

\textbf{Intuitive Explanation.}
Consider a random variable $x$ with a probability distribution $p(x)$ that is supported on a low-dimensional manifold $\mathcal{M} \subset \mathbb{R}^D$. Since the distribution is confined to the manifold, its entire probability measure is concentrated on $\mathcal{M}$. This implies that for any off-manifold point in the ambient space, the score (the gradient of the log data distribution) will point towards the manifold $\mathcal{M}$.

This leads to a key distinction in the behavior of the score field, as illustrated in Fig. \ref{fig:manifold_score}. For an in-distribution point $x \in \mathcal{M}$, the scores in its local ambient neighborhood will converge radially inward toward $\mathcal{M}$. Conversely, for an out-of-distribution (anomalous) point $x \notin \mathcal{M}$, the scores in its neighborhood will align in a coherent direction, pointing towards the manifold 
$\mathcal{M}$.

This distinction in the behavior of the score field forms the basis of our method to determine whether a point is anomalous. A formal proof is provided below.

\textbf{Perturbation Kernel.}
In this work, for a given data point $x$, we use the Gaussian perturbation kernel to generate perturbed points in its neighborhood \cite{song2020score}, \cite{song2019generative}. The process is formally defined as follows:
\begin{equation}
p(\mathbf{x}_{t}|\mathbf{x})=\mathcal{N}(\mathbf{x}_{t}|\mathbf{x},\sigma_{t}^{2}\mathbf{I})
\label{eq:perturbation_kernal}
\end{equation}
Where $\mathbf{x}_{t}$ is the perturbed datapoint of $x$ (represents the neighborhood datapoint of $x$), $p(\mathbf{x}_{t}|\mathbf{x})$ is the probability distribution of the perturbed datapoint, $\sigma_{t}$ is the standard deviation corresponding to perturbation time $t$, $\sigma_{t}=\sqrt{(\sigma^{2t} - 1)/(2\log(\sigma))}$.

\textbf{Theorem 2.} \cite{stanczuk2024diffusion}
\label{thm:score_direction}
\textit{Suppose that the support of the data distribution $p_0$ is contained in a manifold $\mathcal{M} \subset \mathbb{R}^D$ and let $p_t$ be the perturbation distribution of samples from $p_0$ diffused for time $t$, depicted as Eq. \ref{eq:perturbation_kernal}. Then, for any point $x_t \in \mathbb{R}^D$ sufficiently close to $\mathcal{M}$ with orthogonal projection on $\mathcal{M}$ given by $\pi(x_t)$, and $\mathbf{n} = (\pi(x_t) - x_t) / ||\pi(x_t) - x_t||$, we have:
\begin{equation}
S_{\text{cos}}(\mathbf{n}, \nabla_{x_t} \ln p_t(x_t)) \underset{t \to 0}{\longrightarrow} 1
\end{equation}
where $S_{\text{cos}}$ denotes the cosine similarity, defined as $S_{\text{cos}}(\mathbf{a}, \mathbf{b}) = \frac{\mathbf{a} \cdot \mathbf{b}}{||\mathbf{a}|| \cdot ||\mathbf{b}||}$. In other words, for sufficiently small $t$ the score $\nabla_{x_t} \ln p_t(x_t)$ points directly at the projection of $\mathbf{x}$ on the manifold $\mathcal{M}$.}

As a direct corollary of Theorem 2, the score $\nabla_{x_t} \ln p_t(x_t)$ lies in the normal space to the manifold $\mathcal{M}$ at $\pi(x_t)$. This is expressed by the following equation:
\begin{equation}
    \nabla_{x_t} \ln p_t(x_t) \in N_{\pi(x_t)}\mathcal{M}
    \label{eq:score_normal_space}
\end{equation}
Where $N_{\pi(x_t)}\mathcal{M}$ denotes the normal space of the manifold $\mathcal{M}$ at point $\pi(x_t)$.

\textbf{Lemma 2.} \textit{For a point $x$ on the manifold $\mathcal{M}$, let $x_t$ be a sample obtained through a Gaussian perturbation kernel. When $t \to 0$, we have:
\begin{equation}
\nabla_{x_t} \ln p_t(x_t) \in N_{x}\mathcal{M}
\end{equation}
}
\begin{proof}
As $t \to 0$, we have $\pi(x_t) \to x$. Since $\pi(x_t)$ is close to $x$ for a small $t$, the tangent spaces are nearly parallel, which implies that the normal spaces are also nearly identical, i.e., $N_{\pi(x_t)}\mathcal{M} \approx N_x\mathcal{M}$. Furthermore, according to Eq. \ref{eq:score_normal_space}, we have $\nabla_{x_t} \ln p_t(x_t) \in N_{\pi(x_t)}\mathcal{M}$. Combining these yields: $\nabla_{x_t} \ln p_t(x_t) \in N_{x}\mathcal{M} \approx N_{\pi(x_t)}\mathcal{M} $.
\end{proof}

According to Lemma 2, when $t$ is sufficiently small and a sufficient number of perturbed datapoints (K samples) are sampled for $x$ via a Gaussian perturbation kernel, the corresponding scores will approximately span the normal space $N_x\mathcal{M}$, which implies that the rank of the set of $K$ scores is equal to the dimension of the normal space, i.e.:
\begin{equation}
r_{scores} = \text{Dim}(N_{x}\mathcal{M}) = D-d
\end{equation}
Where $D$ is the ambient dimension and $d$ is the intrinsic dimension of the manifold $\mathcal{M}$.

Assume that the intrinsic dimension $d$ of the manifold $\mathcal{M}$ is much smaller than the ambient dimension $D$ ($d \ll D$). Since the $K$ perturbed datapoints are sampled randomly in the ambient space via a Gaussian perturbation kernel, the directions of the scores for almost all these points (except for a negligible few falling on $\mathcal{M}$) will be randomly distributed in a $(D-d)$-dimensional space. This implies that the score directions are uniformly distributed across these $(D-d)$ dimensions. Consequently, when a sufficiently large number of scores ($K$ is large) are normalized and summed, the magnitude of the resulting vector will scale with the square root of $K$, which can be expressed as:
\begin{equation}
\left\lVert S \right\rVert = \left\lVert \sum_{i=1}^K \frac{\nabla_{x_t^{i}} \ln p_t(x_t^{i})}{||\nabla_{x_t^{i}} \ln p_t(x_t^{i})||} \right\rVert \underset{K \to \infty}{\longrightarrow} \sqrt{K}
\label{eq:Sn}
\end{equation}
Furthermore, the expectation of $\left\lVert S \right\rVert$ satisfies:
\begin{equation}
E[\|S\|] = E[\sqrt{\|S\|^2}] \le \sqrt{E[\|S\|^2]} = \sqrt{K}
\end{equation}

In stark contrast, the case is entirely different for a point $y$ located far from the manifold $\mathcal{M}$. Its $K$ perturbed datapoints will likewise be distant from the manifold. Crucially, instead of being randomly oriented, the scores of these points will all consistently point back towards the manifold. Consequently, for a point $y$ sufficiently distant from $\mathcal{M}$, the $K$ normalized score vectors are largely co-directional. When aggregated, the magnitude of their resultant vector, $||S_y||$, will therefore approach $K$, as given by the following equation:
\begin{equation}
\text{dist}(y, \mathcal{M}) \to \infty \implies ||S_y|| \to K
\end{equation}

Due to the significant differences between the scores of perturbed datapoints around datapoints on the manifold and those around datapoints outside the manifold, particularly in the diversity of score directions, we can leverage this to distinguish between points on the manifold and those outside it. 

The specific algorithm process is as follows: first, we will sample $K$ times according to Eq. \ref{eq:perturbation_kernal}, resulting in $K$ perturbed datapoints for each tested datapoint. Then, we will treat these perturbed datapoints as the input of our SGM model to obtain their estimated scores. Next, we will normalize these $K$ score vectors to obtain the corresponding unit vectors. Finally, these unit vectors will be summed to form an anomaly vector, and the magnitude of the anomaly vector will serve as the anomalous degree for each datapoint. 

\subsection{Implementation of the ScoreAD on HAD}
\label{sec:application_on_had}

\begin{algorithm}[!t]
    \footnotesize
    \caption{The ScoreAD Algorithm}
    \begin{itemize}
    \item \textbf{Input:}   hyperspectral image $X$; 
                            perturbation time $t$; 
                            number of the perturbation $K$.
    \item \textbf{Output:} anomaly detection map $A$.
    \end{itemize}
    \begin{algorithmic}[1]

    \State Reshape $X^{H\times W\times C}$ into $D^{N\times C} = \{x_1, x_2,..., x_N\}$
    \State Train the score-based generative model with dataset $D$
    \For{$n = 1$ to $N$}
        \State $S_n = 0$
        \State Extract the contextual spectra $c_n$ for $x_n$
        \For{$i = 1$ to $K$}
            \State Sample ${x}_{n,t}^{i}\sim\mathcal{N}({x}_{n,t}| {x}_{n},\sigma_{t}^{2}\mathbf{I})$ 
            \State Input ${x}_{n,t}^{i}, c_n$ to SGM to get $s_{\theta}({x}_{n,t}^{i},t)$
            \State $S_n = S_n + \frac{s_{\theta}({x}_{n,t}^{i},t)}{||s_{\theta}({x}_{n,t}^{i},t)||}$
        \EndFor 
        \State $A_n = ||S_n||$
    \EndFor 
    \State Reshape $A$ from size $N$ into size of $H\times W$

    \end{algorithmic}
    \label{algm.HAD}
\end{algorithm}

As mentioned above, the background spectra (normal spectra) in hyperspectral images also conform to the manifold hypothesis, i.e., they lie on a low-dimensional manifold embedded in the high-dimensional ambient space. Therefore, the anomaly detection method described previously, which utilizes the distributional differences between normal and anomalous data in high-dimensional space, is well applicable to hyperspectral anomaly detection.

As required by Eq. \ref{eq:Sn}, our method needs access to the score of the data distribution $p(x)$. However, the analytical form of $p(x)$ is unknown, making the true score intractable. To overcome this challenge, we employ a score-based generative model (SGM) to estimate this score function. The SGM is trained on the set of all spectra obtained by reshaping an $H \times W \times C$ hyperspectral image into $H \times W$ $C$-dimensional vectors. Once trained, the SGM provides an approximation of the true score, formally expressed as:
\begin{equation}
s_{\theta}(x_t, t) \approx \nabla_{x_t} \ln p_t(x_t)
\end{equation}
Where $s_{\theta}(x_t, t)$ is the score of the data distribution at time $t$, $\theta$ represents the parameters of the SGM, and $p_t(x)$ is the perturbed data distribution of spectra within HSI.

A key limitation of this approach, however, is that it treats each spectrum independently when training the SGM. This completely neglects the spatial context of the hyperspectral image, leading to a loss of information. To alleviate this limitation, we incorporate spatial information into the SGM via a conditional generation framework.

\begin{figure}[!h]
\centering
\includegraphics[width=\linewidth]{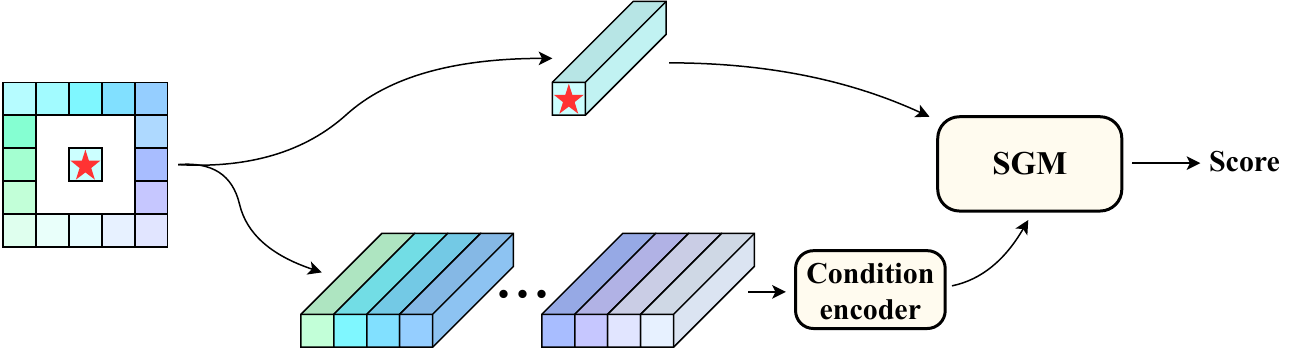}
\caption{
    Overall architecture of ScoreAD. The spectrum marked with red star represents the test spectrum, and the spectra around test spectrum are the contextual spectra within a dual-window neighborhood. The test spectrum is first perturbed via a Gaussian perturbation kernel to obtain a perturbed spectrum, which is then fed into the score-based generative model (SGM) along with its contextual spectra to estimate the score.
}
\label{fig:overall_architecture}
\end{figure}

More precisely, for every input sample (spectrum), we extract its contextual spectra within a dual-window neighborhood. These neighboring spectra then serve as conditional inputs to a conditional encoder, which modulates the model's output accordingly. This process is illustrated in Fig. \ref{fig:overall_architecture}. Our implementation of the score-based generative model (SGM) employs 1D convolutions to capture inter-spectral correlations and utilizes the FiLM \cite{FiLM} mechanism to inject these spatial conditions.

First, we train our score-based generative model using all spectra from the hyperspectral image. Once trained, the model is used to perform K inferences for each spectrum, yielding K corresponding Scores. These K Scores are subsequently normalized and summed according to Eq. \ref{eq:Sn} to yield $S_n$. The magnitude of $S_n$ is then used as the anomaly degree for each spectrum. The detailed steps are shown in Algorithm \ref{algm.HAD}.

\section{Experiments and Analysis}
In this section, we will evaluate the detection performance of ScoreAD on four hyperspectral datasets.

\subsection{Experimental Setup}

\subsubsection{Datasets} 
We evaluate our proposed method both on four hyperspectral datasets, including HYDICE, Pavia, Hyperion and Salinas. HYDICE dataset consists of 100$\times$80 pixels and 162 spectral bands in a wavelength range of 400 to 2500nm. Pavia dataset is a 150$\times$150$\times$102 hyperspectral image, with the wavelength range from 430 to 860nm. Hyperion dataset is a 100$\times$100 image, with 145 spectral bands. Salinas dataset is a 120$\times$120$\times$204 hyperspectral image. 

\subsubsection{Comparison Methods} 
To fully demonstrate the effectiveness of ScoreAD, we selected ten classic and effective anomaly detection algorithms as comparison methods, including RXAD~\cite{RXAD}, CRD~\cite{CRD}, LSMAD~\cite{LSMAD}, PTA~\cite{PTA}, AutoAD~\cite{Auto-AD}, RGAE~\cite{RGAE}, BSDM~\cite{BSDM}, TAEF~\cite{TAEF}, GT-HAD~\cite{GT-HAD}. These methods cover the three mainstream categories of HAD. RXAD is the classical statistics-based HAD method, CRD, LSMAD and PTA are the classical representation-based HAD method, while AutoAD, RGAE, BSDM, TAEF, GT-HAD are belong to the class of deep learning-based HAD method. Among these,  BSDM is a method based on diffusion models that uses diffusion to denoise the background in hyperspectral images, aiming to enhance the distinction between the background and anomalies.

\subsubsection{Metrics}
For a comprehensive evaluation of our hyperspectral anomaly detection performance, we employ $\mathrm{AUC}_{\mathrm{PR}}$ metric and 3D-ROC derived metrics~\cite{3DROC}, including $\text{AUC}_\text{(D,F)}$, $\text{AUC}_{(\text{D},\tau)}$, $\text{AUC}_{(\text{F},\tau)}$, $\text{AUC}_\text{TD}$, $\text{AUC}_\text{BS}$, $\text{AUC}_\text{SNPR}$, $\text{AUC}_\text{TD-BS}$ and $\text{AUC}_\text{ODP}$. The $\mathrm{AUC}_{\mathrm{PR}}$ metric is particularly suited for addressing the class imbalance problem often present in anomaly detection, thus providing a more robust performance assessment on such datasets. In addition, box-whisker plot is also used for our evaluation, which visually represents the degree of separation between anomalous and normal data~\cite{box_plot}.
For a comprehensive evaluation of our hyperspectral anomaly detection performance, we employed the 3D-ROC (three dimensional receiver operating characteristic) curve and its derived eight metrics to assess the overall effectiveness of the detection algorithms, as well as their anomaly detectability and background suppression capabilities \cite{3D-ROC-TD}, \cite{3D-ROC-AD}.
The eight derived metrics of 3D-ROC curve are
$\text{AUC}_\text{(D,F)}$, $\text{AUC}_{(\text{D},\tau)}$, $\text{AUC}_{(\text{F},\tau)}$, $\text{AUC}_\text{TD}$, $\text{AUC}_\text{BS}$, $\text{AUC}_\text{SNPR}$, $\text{AUC}_\text{TD-BS}$ and $\text{AUC}_\text{ODP}$. 

\begin{figure*}[!ht]
\centering
\includegraphics[width=\linewidth]{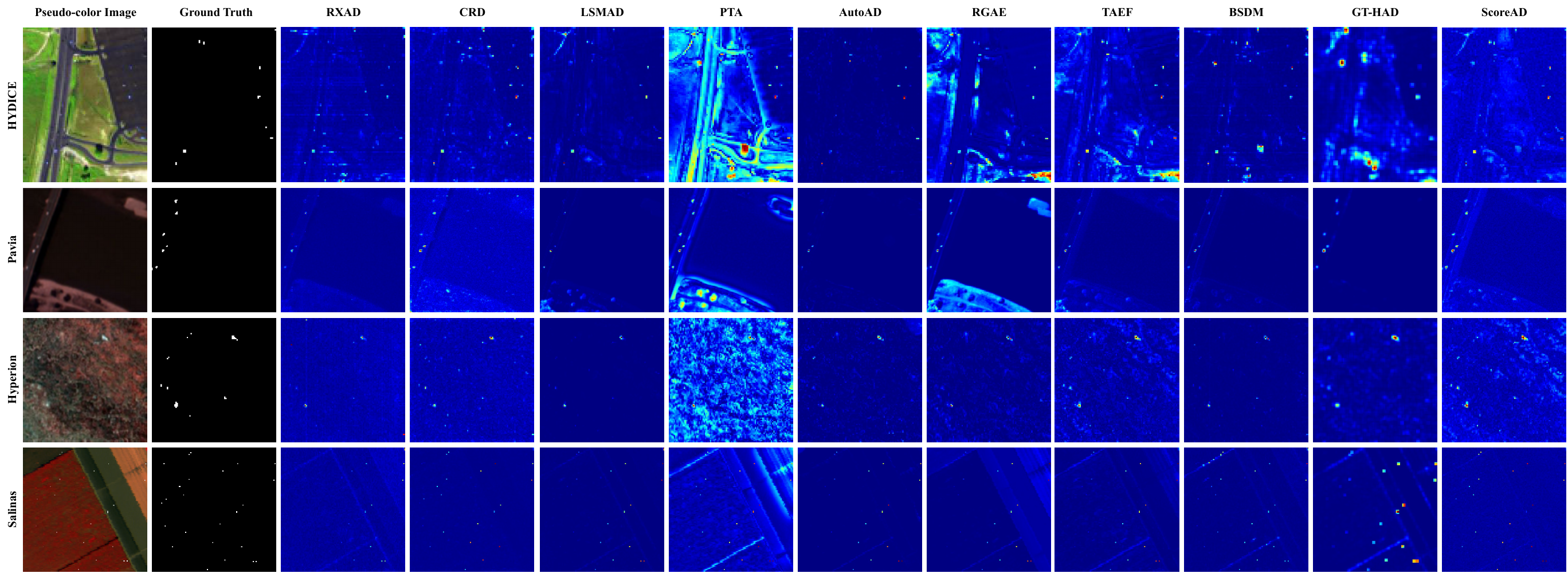}
\caption{
    The pseudo-color image, Ground Truth, and hyperspectral anomaly detection maps of the comparison methods on four hyperspectral datasets.
}
\label{fig:detection_map}
\end{figure*}

\begin{figure}[!ht]
    \centering
    
    \begin{subfigure}{0.45\columnwidth}
        \centering
        \includegraphics[width=\linewidth]{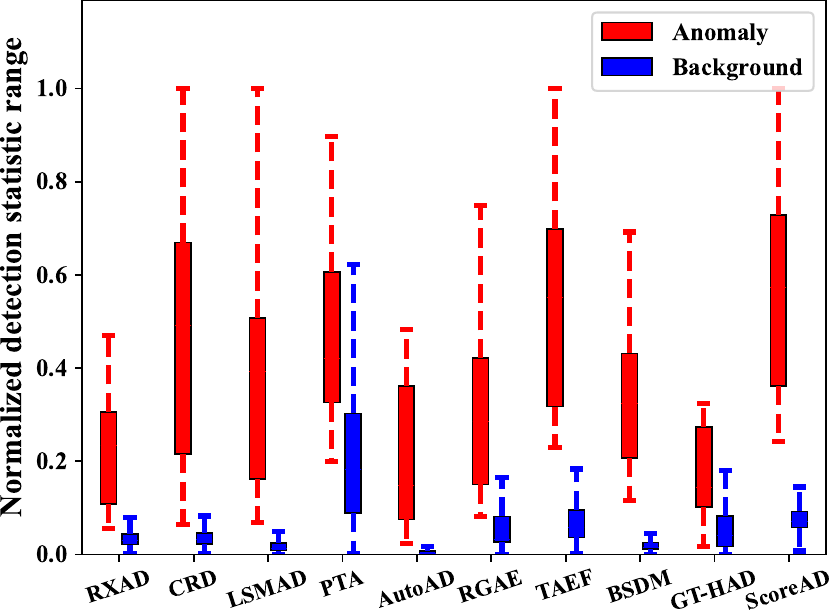}
        \caption{HYDICE}
        \label{fig:HYDICE_box}
    \end{subfigure}\hfill
    \begin{subfigure}{0.45\columnwidth}
        \centering
        \includegraphics[width=\linewidth]{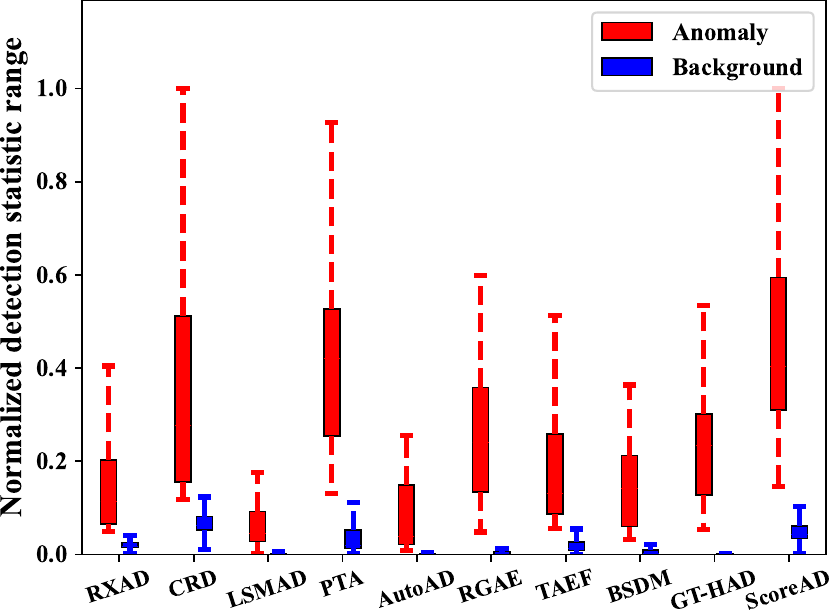}
        \caption{Pavia}
        \label{fig:Pavia_box}
    \end{subfigure}
    
    \vspace{0.1cm}
    
    \begin{subfigure}{0.45\columnwidth}
        \centering
        \includegraphics[width=\linewidth]{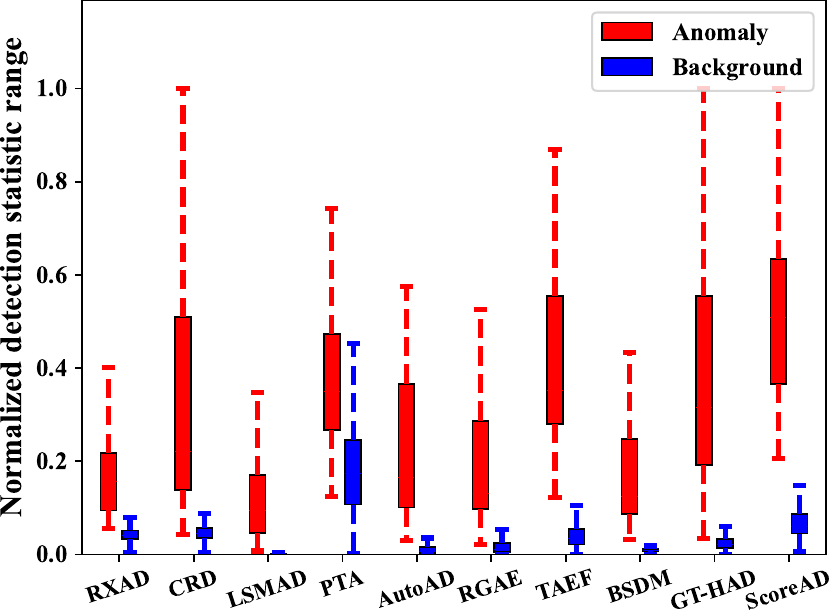}
        \caption{Hyperion}
        \label{fig:Hyperion_box}
    \end{subfigure}\hfill
    \begin{subfigure}{0.45\columnwidth}
        \centering
        \includegraphics[width=\linewidth]{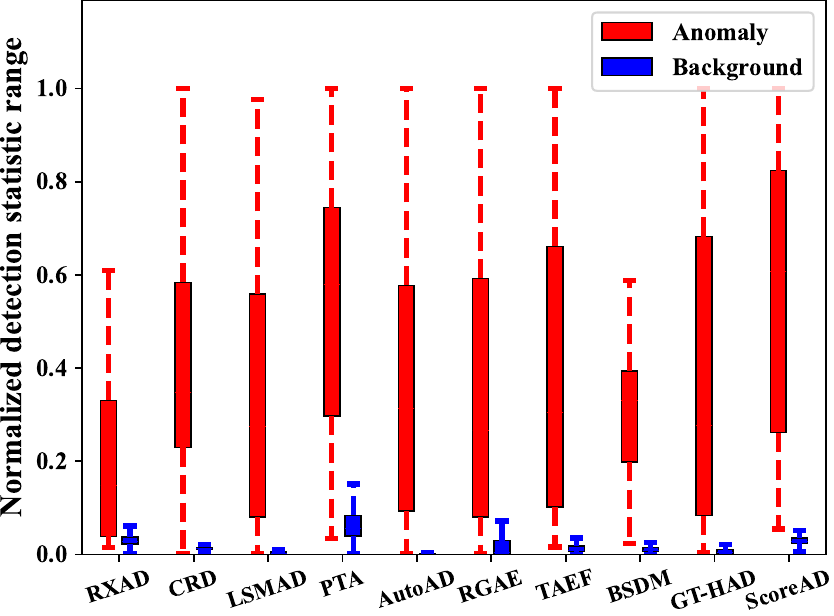}
        \caption{Salinas}
        \label{fig:Salinas1_box}
    \end{subfigure}
    
    \caption{Box-whisker plots of different methods on four datasets. Greater separability between the background and anomalies indicates better detection performance.}
    \label{fig:overall_box_plots}
\end{figure}

\subsubsection{Implementation Details}
To facilitate a fair and robust comparison of method effectiveness, we standardized the parameter settings for all algorithms, including our own, across the four datasets. These settings were determined based on the original papers and associated code. The specific configurations are as follows:

\begin{itemize}
    \item \textbf{RXAD:} Parameter-free.
    
    \item \textbf{CRD:} The parameters ($w_{\text{in}}, w_{\text{out}}, \lambda$) were set to ($5, 7, 10^{-6}$).
    
    \item \textbf{LSMAD:} The number of iterations \texttt{Iter} and $\epsilon$ were set to $10^2$ and $10^{-3}$.
    
    \item \textbf{PTA:} The parameters ($r, \alpha, \tau, \beta$) were set to ($1, 1, 1, 0.1$).
    
    \item \textbf{Auto-AD:} The parameter $\sigma$ was set to $1.5 \times 10^{-5}$.
    
    \item \textbf{RGAE:} The parameters ($\lambda, S$) were set to ($10^{-3}, 150$).
    
    \item \textbf{TAEF:} The number of heads, $N_{\text{heads}}$, was set to 6 for the HYDICE, Pavia and Salinas datasets, and 5 for the Hyperion dataset.
    
    \item \textbf{BSDM:} The parameter $\lambda$ was set to $10^{-4}$.
    
    \item \textbf{GT-HAD:} The parameter $\lambda$ was set to $10^{-3}$.
    
    \item \textbf{ScoreAD:} The parameter $(w_{\text{in}}, w_{\text{out}})$ was set to $(3, 5)$ for Pavia, Hyperion and Salinas datasets. The perturbation time $t$ was set to $0.05$ for the HYDICE, Pavia and Salinas datasets, and $0.01$ for the Hyperion. The number of perturbations $K$ was set to $100$ for all datasets.
\end{itemize}

All compared methods were implemented based on their official code or original papers. For each dataset, we empirically tuned the parameters of the respective methods, with primarily parameter settings provided above. All experiments were conducted on a computer equipped with an Intel i7-13700K CPU and an NVIDIA RTX 4060Ti GPU. Among these, the non-deep learning methods and RGAE were implemented in MATLAB R2023a.

\begin{figure*}[!ht]
\centering
\includegraphics[width=\linewidth]{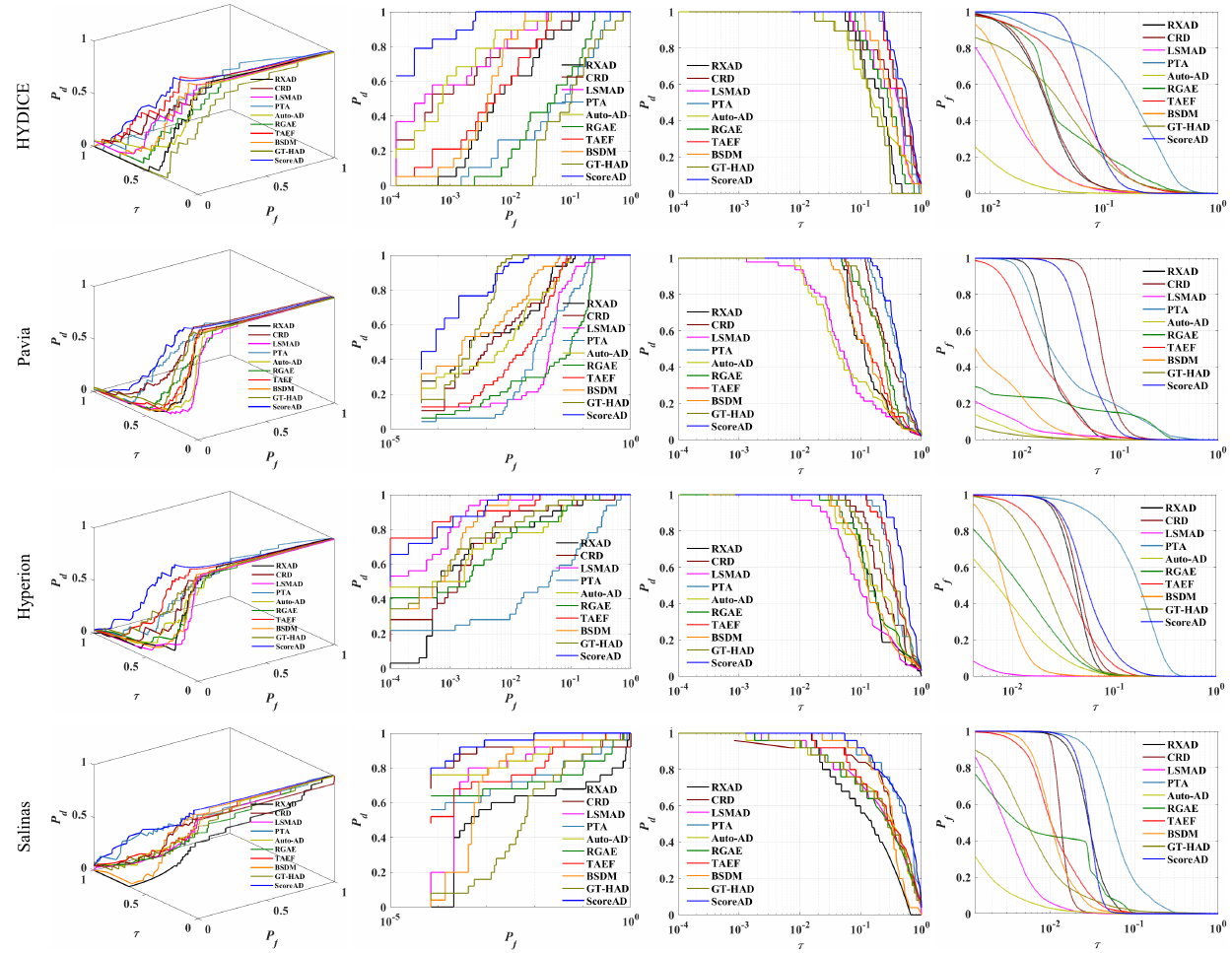}
\caption{
    3D-ROC curves of different HAD methods on four datasets.
}
\label{fig:roc}
\end{figure*}

\begin{figure*}[!ht]
    \centering
    
    \begin{subfigure}{0.48\columnwidth}
        \centering
        \includegraphics[width=\linewidth]{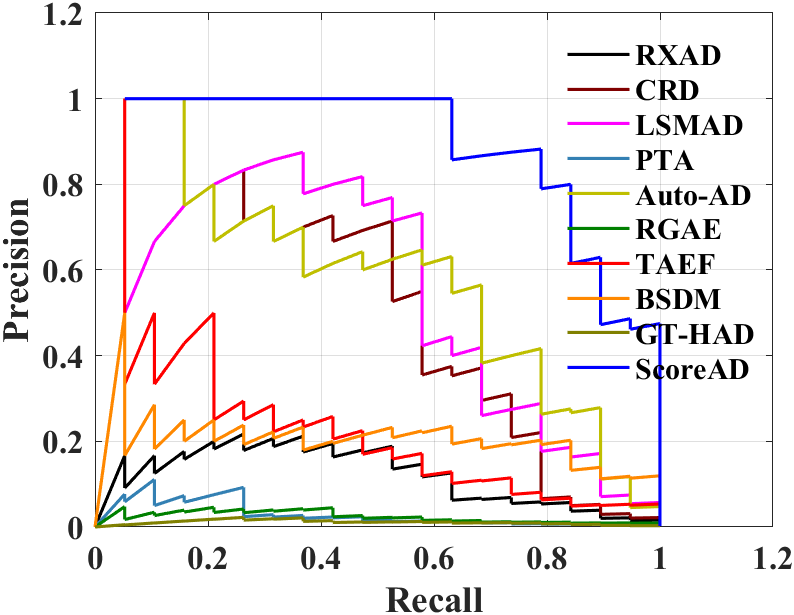}
        \caption{HYDICE}
    \end{subfigure}\hfill
    \begin{subfigure}{0.48\columnwidth}
        \centering
        \includegraphics[width=\linewidth]{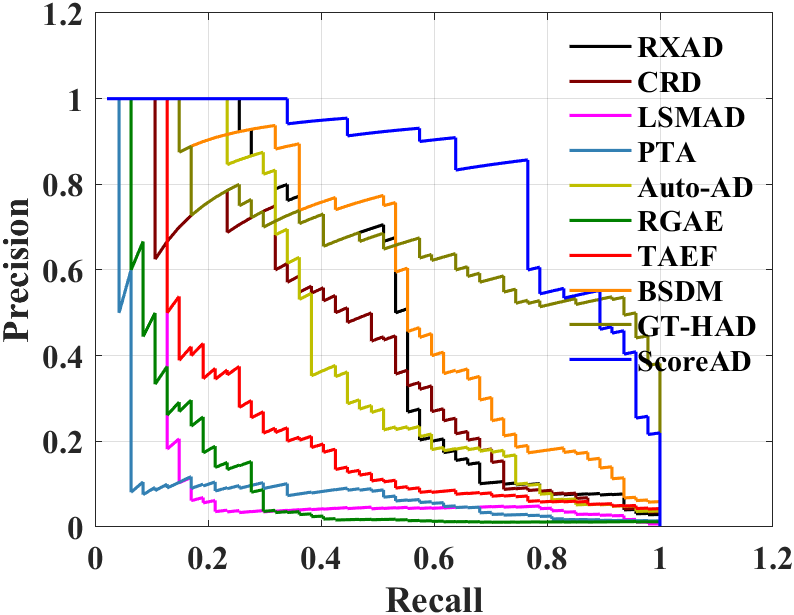}
        \caption{Pavia}
    \end{subfigure}\hfill
    \begin{subfigure}{0.48\columnwidth}
        \centering
        \includegraphics[width=\linewidth]{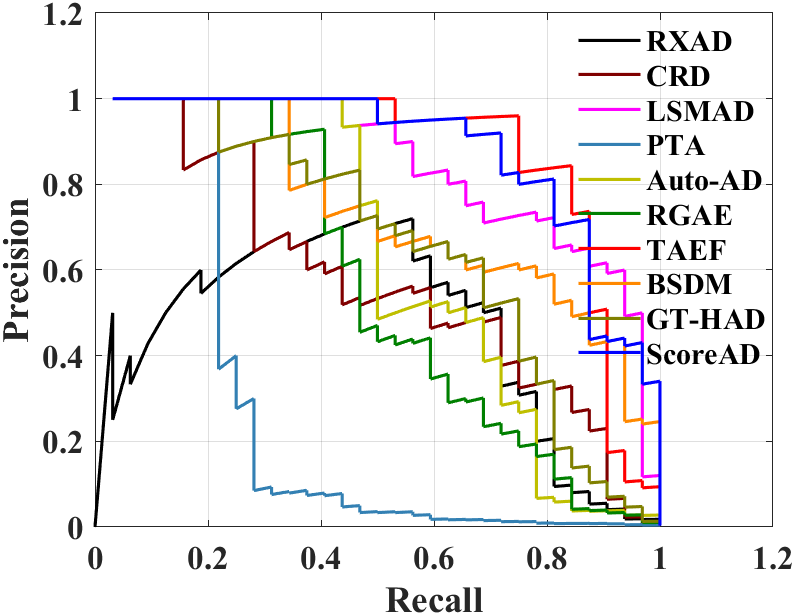}
        \caption{Hyperion}
    \end{subfigure}\hfill
    \begin{subfigure}{0.48\columnwidth}
        \centering
        \includegraphics[width=\linewidth]{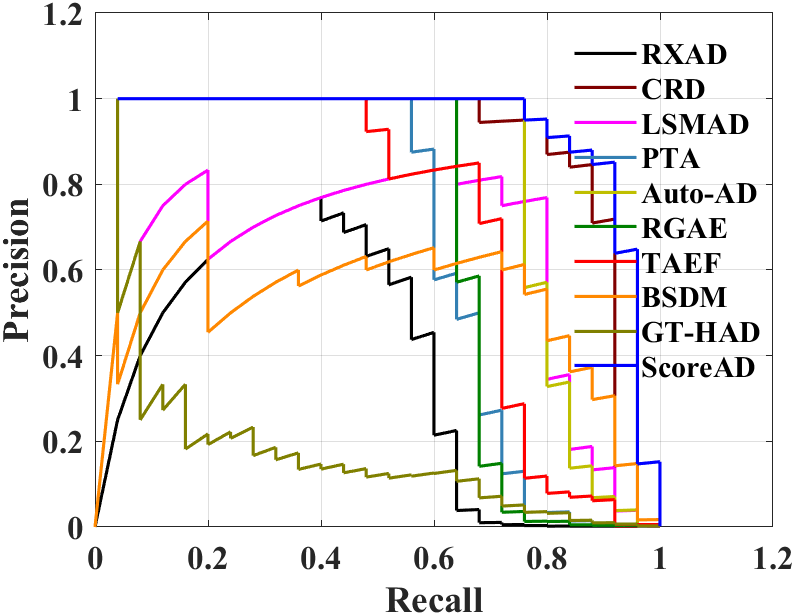}
        \caption{Salinas}
    \end{subfigure}\hfill
    
    \caption{PR curves of different HAD methods on four datasets.}
    \label{fig:PR curve}
\end{figure*}

\begin{table*}[!h]
    \centering
    \caption{3D-ROC metric values of different methods across four hyperspectral datasets. The best result is in \textbf{bold} and the second-best is \underline{underlined}.}
    \label{tab:3D-ROC metrics}
    \begin{tabular}{llcccccccccc}
    \toprule
    \textbf{Dataset} & \textbf{Metric} & \textbf{RXAD} & \textbf{CRD} & \textbf{LSMAD} & \textbf{PTA} & \textbf{AutoAD} & \textbf{RGAE} & \textbf{TAEF} & \textbf{BSDM} & \textbf{GT-HAD} & \textbf{ScoreAD} \\
    \midrule
    \multirow{8}{*}{HYDICE} 
    & $\text{AUC}_{(P_d,P_f)}$ & 0.9763 & 0.9855 & 0.9943 & 0.8715 & \underline{0.9952} & 0.9143 & 0.9870 & 0.9943 & 0.8358 & \textbf{0.9996} \\
    & $\text{AUC}_{(P_d,\tau)}$ & 0.2198 & 0.4331 & 0.3604 & 0.4780 & 0.2790 & 0.3074 & \underline{0.5135} & 0.3656 & 0.1734 & \textbf{0.5660} \\
    & $\text{AUC}_{(P_f,\tau)}$ & 0.0381 & 0.0391 & \underline{0.0213} & 0.2062 & \textbf{0.0066} & 0.0767 & 0.0795 & 0.0250 & 0.0673 & 0.0792 \\
    & $\text{AUC}_{\text{TD}}$ & 1.1961 & 1.4187 & 1.3548 & 1.3496 & 1.2743 & 1.2217 & \underline{1.5005} & 1.3599 & 1.0091 & \textbf{1.5656} \\
    & $\text{AUC}_{\text{BS}}$ & 0.9382 & 0.9464 & \underline{0.9730} & 0.6653 & \textbf{0.9886} & 0.8376 & 0.9075 & 0.9693 & 0.7684 & 0.9204 \\
    & $\text{AUC}_{\text{SNPR}}$ & 5.7681 & 11.0765 & \underline{16.9152} & 2.3185 & \textbf{41.9719} & 4.0096 & 6.4607 & 14.6443 & 2.5761 & 7.1471 \\
    & $\text{AUC}_{\text{TD-BS}}$ & 0.1817 & 0.3940 & 0.3391 & 0.2719 & 0.2724 & 0.2307 & \underline{0.4340} & 0.3407 & 0.1061 & \textbf{0.4868} \\
    & $\text{AUC}_{\text{ODP}}$ & 1.1580 & 1.3796 & 1.3334 & 1.1434 & 1.2676 & 1.1450 & \underline{1.4210} & 1.3349 & 0.9418 & \textbf{1.4864} \\
    \midrule
    \multirow{8}{*}{Pavia} 
    & $\text{AUC}_{(P_d,P_f)}$ & 0.9905 & 0.9905 & 0.9653 & 0.9635 & 0.9900 & 0.9277 & 0.9869 & 0.9956 & \textbf{0.9992} & \underline{0.9987} \\
    & $\text{AUC}_{(P_d,\tau)}$ & 0.1730 & 0.3567 & 0.1203 & \textbf{0.3952} & 0.1380 & 0.2653 & 0.2087 & 0.1830 & 0.2828 & \underline{0.3880} \\
    & $\text{AUC}_{(P_f,\tau)}$ & 0.0233 & 0.0705 & 0.0052 & 0.0590 & \underline{0.0019} & 0.0375 & 0.0213 & 0.0075 & \textbf{0.0017} & 0.0499 \\
    & $\text{AUC}_{\text{TD}}$ & 1.1635 & 1.3472 & 1.0856 & \underline{1.3588} & 1.1280 & 1.1930 & 1.1955 & 1.1786 & 1.2820 & \textbf{1.3867} \\
    & $\text{AUC}_{\text{BS}}$ & 0.9672 & 0.9200 & 0.9601 & 0.9045 & \underline{0.9881} & 0.8902 & 0.9656 & \underline{0.9881} & \textbf{0.9975} & 0.9488 \\
    & $\text{AUC}_{\text{SNPR}}$ & 7.4149 & 5.0577 & 23.0075 & 6.7001 & \underline{72.5371} & 7.0672 & 9.8062 & 24.3738 & \textbf{168.1419} & 7.7812 \\
    & $\text{AUC}_{\text{TD-BS}}$ & 0.1497 & 0.2861 & 0.1150 & \underline{0.3362} & 0.1361 & 0.2277 & 0.1874 & 0.1755 & 0.2811 & \textbf{0.3382} \\
    & $\text{AUC}_{\text{ODP}}$ & 1.1402 & 1.2767 & 1.0803 & \underline{1.2998} & 1.1261 & 1.1555 & 1.1743 & 1.1711 & 1.2803 & \textbf{1.3369} \\
    \midrule
    \multirow{8}{*}{Hyperion} 
    & $\text{AUC}_{(P_d,P_f)}$ & 0.9829 & 0.9753 & \textbf{0.9987} & 0.8676 & 0.9826 & 0.9771 & 0.9975 & \underline{0.9985} & 0.9868 & \underline{0.9985} \\
    & $\text{AUC}_{(P_d,\tau)}$ & 0.2319 & 0.2996 & 0.1745 & 0.4063 & 0.2627 & 0.2479 & \underline{0.4261} & 0.2170 & 0.3857 & \textbf{0.5717} \\
    & $\text{AUC}_{(P_f,\tau)}$ & 0.0435 & 0.0480 & \textbf{0.0018} & 0.1802 & 0.0130 & 0.0183 & 0.0427 & \underline{0.0096} & 0.0267 & 0.0827 \\
    & $\text{AUC}_{\text{TD}}$ & 1.2148 & 1.2749 & 1.1733 & 1.2739 & 1.2453 & 1.2250 & \underline{1.4236} & 1.2155 & 1.3724 & \textbf{1.5702} \\
    & $\text{AUC}_{\text{BS}}$ & 0.9395 & 0.9273 & \textbf{0.9970} & 0.6874 & 0.9696 & 0.9588 & 0.9548 & \underline{0.9889} & 0.9601 & 0.9158 \\
    & $\text{AUC}_{\text{SNPR}}$ & 5.3358 & 6.2409 & \textbf{99.1037} & 2.2542 & 20.1910 & 13.5587 & 9.9706 & \underline{22.6681} & 14.4513 & 6.9150 \\
    & $\text{AUC}_{\text{TD-BS}}$ & 0.1884 & 0.2516 & 0.1728 & 0.2260 & 0.2497 & 0.2296 & \underline{0.3833} & 0.2075 & 0.3590 & \textbf{0.4890} \\
    & $\text{AUC}_{\text{ODP}}$ & 1.1714 & 1.2269 & 1.1715 & 1.0937 & 1.2323 & 1.2067 & \underline{1.3809} & 1.2059 & 1.3458 & \textbf{1.4875} \\
    \midrule
    \multirow{8}{*}{Salinas} 
    & $\text{AUC}_{(P_d,P_f)}$ & 0.8073 & 0.9166 & 0.9733 & 0.9294 & 0.9841 & 0.9144 & 0.9737 & \underline{0.9950} & 0.9423 & \textbf{0.9987} \\
    & $\text{AUC}_{(P_d,\tau)}$ & 0.2242 & 0.4130 & 0.3843 & \underline{0.5395} & 0.3837 & 0.3590 & 0.4017 & 0.3490 & 0.3531 & \textbf{0.5673} \\
    & $\text{AUC}_{(P_f,\tau)}$ & 0.0314 & 0.0140 & \underline{0.0042} & 0.0731 & \textbf{0.0019} & 0.0189 & 0.0150 & 0.0133 & 0.0143 & 0.0314 \\
    & $\text{AUC}_{\text{TD}}$ & 1.0315 & 1.3297 & 1.3576 & \underline{1.4689} & 1.3679 & 1.2734 & 1.3754 & 1.3440 & 1.2954 & \textbf{1.5660} \\
    & $\text{AUC}_{\text{BS}}$ & 0.7759 & 0.9026 & 0.9691 & 0.8563 & \textbf{0.9822} & 0.8955 & 0.9588 & \underline{0.9817} & 0.9280 & 0.9673 \\
    & $\text{AUC}_{\text{SNPR}}$ & 7.1390 & 29.5495 & \underline{91.5942} & 7.3834 & \textbf{199.9108} & 18.9882 & 26.8102 & 26.2580 & 24.6726 & 18.0615 \\
    & $\text{AUC}_{\text{TD-BS}}$ & 0.1928 & 0.3991 & 0.3801 & \underline{0.4664} & 0.3818 & 0.3401 & 0.3867 & 0.3358 & 0.3388 & \textbf{0.5359} \\
    & $\text{AUC}_{\text{ODP}}$ & 1.0001 & 1.3157 & 1.3534 & \underline{1.3958} & 1.3660 & 1.2545 & 1.3604 & 1.3307 & 1.2810 & \textbf{1.5346} \\
    \bottomrule
    \end{tabular}
\end{table*}

\begin{table*}[!ht]
    \small
    \centering
    \caption{$\text{AUC}_{\text{PR}}$ values of different methods on four hyperspectral datasets and their average. The best result is in \textbf{bold} and the second-best is \underline{underlined}.}
    \label{tab:auc_pr_metric}
    \begin{tabular}{lcccccccccc}
        \toprule
        \textbf{Dataset} & \textbf{RXAD} & \textbf{CRD} & \textbf{LSMAD} & \textbf{PTA} & \textbf{AutoAD} & \textbf{RGAE} & \textbf{TAEF} & \textbf{BSDM} & \textbf{GT-HAD} & \textbf{ScoreAD} \\
        \midrule
        HYDICE   & 0.1186 & 0.4395 & 0.5024 & 0.0276 & \underline{0.5283} & 0.0232 & 0.1832 & 0.1988 & 0.0108 & \textbf{0.8409} \\
        Pavia    & 0.5005 & 0.4159 & 0.1458 & 0.0900 & 0.4274 & 0.1171 & 0.2400 & 0.5651 & \underline{0.6649} & \textbf{0.7375} \\
        Hyperion & 0.4512 & 0.5394 & 0.8105 & 0.2312 & 0.5927 & 0.5270 & \textbf{0.8373} & 0.7024 & 0.6220 & \underline{0.8264} \\
        Salinas  & 0.3578 & \underline{0.8511} & 0.6154 & 0.6179 & 0.7661 & 0.6320 & 0.6638 & 0.5060 & 0.1352 & \textbf{0.8470} \\
        \midrule
        Average  & 0.3570 & 0.5615 & 0.5185 & 0.2417 & \underline{0.5786} & 0.3248 & 0.4811 & 0.4931 & 0.3582 & \textbf{0.8130} \\
        \bottomrule
    \end{tabular}
\end{table*}

\subsection{Detection Performance}

In this section, we will evaluate the detection performance of ScoreAD from both qualitative and quantitative aspects.
The qualitative assessment is conducted via detection maps and box plots, which offer an intuitive visualization of the performance. For the quantitative evaluation, we utilize 3D-ROC and AUC-PR metrics to rigorously measure detection capabilities. Based on these qualitative and quantitative results, we will then provide a detailed discussion of the method's detection performance across the four datasets.

\subsubsection{Performance over HYDICE}
The detection maps over the HYDICE dataset is presented in the first row of Fig. \ref{fig:detection_map}. From the detection map, we can see that ScoreAD successfully detected all anomalous pixels without any missed detections, and its background suppression capability is also commendable. Correspondingly, as shown in Table \ref{tab:3D-ROC metrics} and \ref{tab:auc_pr_metric}, the metrics for ScoreAD reflect this performance, with $\text{AUC}_{(P_d,P_f)}$, $\text{AUC}_{(P_d,\tau)}$, $\text{AUC}_{\text{TD}}$, $\text{AUC}_{\text{TD-BS}}$, $\text{AUC}_\text{ODP}$ and $\text{AUC}_{\text{PR}}$ being best, indicating a solid overall detection performance and excellent target detectability, which aligns with the observations from the detection map and box-whisker plot.

\subsubsection{Performance over Pavia}
The detection maps and box-whisker plot over the Pavia dataset are shown in Fig. \ref{fig:detection_map} and Fig. \ref{fig:Pavia_box}. 
ScoreAD demonstrates excellent object detection and separation capabilities. While its object detection performance is comparable to that of GT-HAD, it is outperformed by GT-HAD in terms of background suppression. This is attributed to the differing design principles of the two methods.

The 3D-ROC and PR curves are presented in Fig. \ref{fig:roc} and Fig. \ref{fig:PR curve}. 
And the corresponding metrics are listed in Table \ref{tab:3D-ROC metrics} and \ref{tab:auc_pr_metric}. 
ScoreAD achieves the best/second-best performance in $\text{AUC}_{(P_d,P_f)}$, $\text{AUC}_{(P_d,\tau)}$, $\text{AUC}_{\text{TD}}$, $\text{AUC}_{\text{TD-BS}}$, $\text{AUC}_{\text{ODP}}$ and $\text{AUC}_{\text{PR}}$, indicating its strong overall detection capability and target detection ability.

\subsubsection{Performance over Hyperion}
The anomaly detection maps over the Hyperion dataset is presented in the third row of Fig. \ref{fig:detection_map}. And the box-whisker plot is shown in Fig. \ref{fig:Hyperion_box}.
Similar to the previous two datasets, ScoreAD also exhibits strong overall detection capabilities in the detection map for the Hyperion dataset.

The ROC and PR curves are displayed in Fig. \ref{fig:roc} and Fig. \ref{fig:PR curve}. The derived metrics are listed in Table \ref{tab:3D-ROC metrics} and \ref{tab:auc_pr_metric}. 
The metrics of $\text{AUC}_{(P_d,\tau)}$, $\text{AUC}_{\text{TD}}$, $\text{AUC}_{\text{TD-BS}}$, $\text{AUC}_{\text{ODP}}$ obtain best results, while $\text{AUC}_{(P_d,P_f)}$, $\text{AUC}_{\text{PR}}$ achieve second-best performance, demonstrating the outstanding overall detection capability and target detection ability of ScoreAD.

\subsubsection{Performance over Salinas}
The detection maps and box plot over the Hyperion dataset are presented in Fig. \ref{fig:detection_map} and Fig. \ref{fig:Salinas1_box}. 
On the Salinas dataset, ScoreAD achieves superior performance in both object detection and background suppression, surpassing the majority of deep learning approaches. Notably, its effectiveness is on par with CRD, a method based on domain collaborative representation.

The curves are displayed in Fig. \ref{fig:roc}, Fig. \ref{fig:PR curve} and the derived metrics are listed in Table \ref{tab:3D-ROC metrics} and \ref{tab:auc_pr_metric}. 
The metrics of $\text{AUC}_{(P_d,P_f)}$, $\text{AUC}_{(P_d,\tau)}$, $\text{AUC}_{\text{TD}}$, $\text{AUC}_{\text{TD-BS}}$, $\text{AUC}_{\text{ODP}}$ and $\text{AUC}_{\text{PR}}$ achieve best results, demonstrating the outstanding overall detection capability and target detection ability of ScoreAD.

Overall, ScoreAD consistently demonstrates strong performance across all four datasets, achieving the best or second-best results in the majority of evaluation metrics. This consistent performance underscores the robustness and effectiveness of ScoreAD in hyperspectral anomaly detection tasks.



\begin{figure}[!h]
\centering
\includegraphics[width=0.9\linewidth]{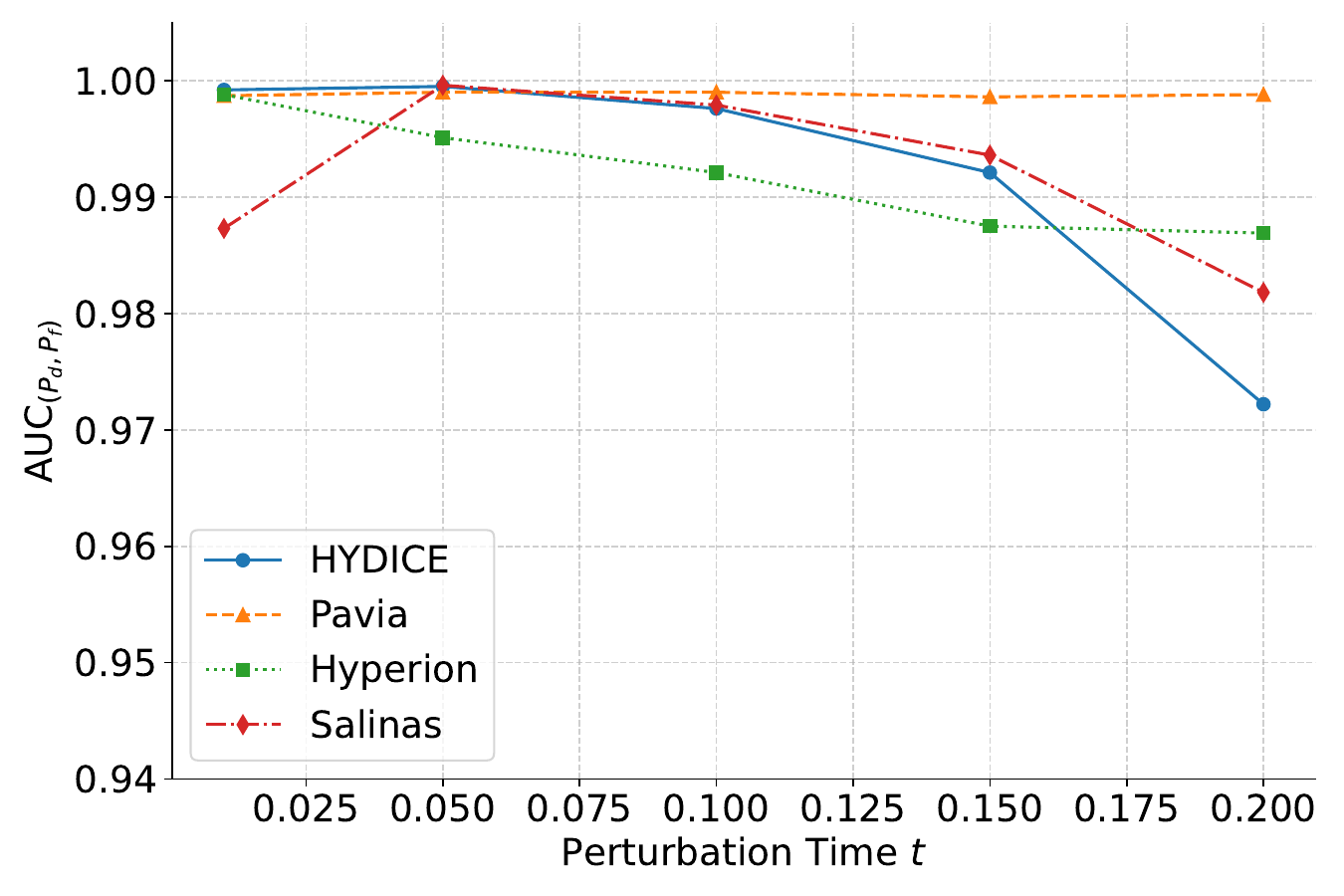}
\caption{
    The Effect of the Perturbation Time on Detection performance ($\text{AUC}_{(P_d,P_f)}$). We evaluated our method across five perturbation time settings: 0.01, 0.05, 0.1, 0.15 and 0.2.
}
\label{fig:ablation_plot}
\end{figure}

\begin{figure}[!h]
\centering
\includegraphics[width=0.92\linewidth]{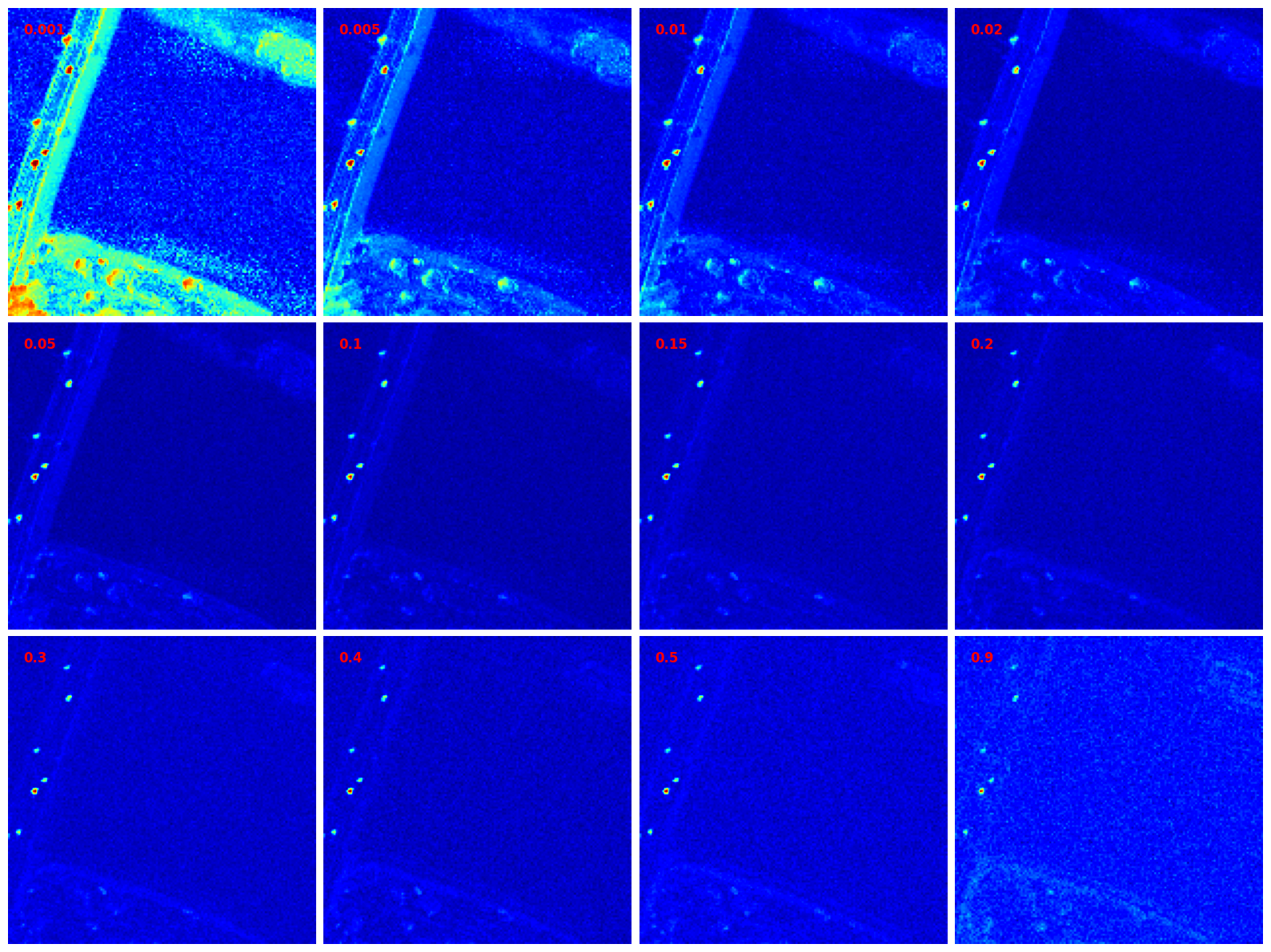}
\caption{
    Detection Maps of ScoreAD on the Pavia Dataset for Different Values of $t$.
}
\label{fig:Pavia_detection_map_12}
\end{figure}

\subsection{Parameter Analysis}

In this subsection, we conducted an ablation study to evaluate the contribution of two key elements in ScoreAD: the perturbation time (parameter $t$) and the conditional input of neighborhood spectra.

\subsubsection{Effect of the Perturbation Time $t$.}
To investigate the effect of the perturbation time, we conduct experiments with different perturbation time ranging from 0.01 to 0.2. The results are shown in fig. \ref{fig:ablation_plot}. It can be observed that for lower values of the perturbation time (0.01, 0.05, 0.1), the performance of our method is relatively stable and strong. In contrast, when a larger perturbation time is used (0.2), the performance of ScoreAD degrades significantly. This finding is consistent with the underlying principle of ScoreAD, which assumes a small perturbation time. Additionally, we found that for $t$=0.05, the results are strong across all four datasets, suggesting that the perturbation time parameter is not highly sensitive to the specific dataset..

\begin{table}[!ht]
    \centering
    \small
    \caption{$\text{AUC}_{(P_d,P_f)}$ of ScoreAD on four hyperspectral datasets with different size of Dual-Window. $\backslash$ indicates the situation without conditional input.}
    \begin{tabular}{cccccc}
        \toprule
        $(W_{in},W_{out})$ & HYDICE & Pavia & Hyperion & Salinas \\
        \midrule
    
        $\backslash$ & 0.9996 & 0.9975 & 0.9989 & 0.9655 \\ 
        $(1,3)$ & 0.9971 & 0.9972 & 0.9982 & 0.9995 \\ 
        $(3,5)$ & 0.9971 & 0.9987 & 0.9985 & 0,9987 \\
        $(1,5)$ & 0.9980 & 0.9979 & 0.9985 & 0.9990 \\
        $(5,7)$ & 0.9955 & 0.9993 & 0.9987 & 0.9979 \\
        \bottomrule
    \end{tabular}
    \label{tab:ablation}
\end{table}

\subsubsection{Effect of Conditional Input.}
We conducted a sensitivity analysis on the dual-window size to assess its impact on model performance. The results are summarized in Table \ref{tab:ablation}. The results indicate that the performance of ScoreAD is not highly sensitive to the dual-window size within a certain range. Moreover, it is evident that for the Salinas dataset, incorporating the conditional input leads to a substantial improvement in the model's detection performance.



\section{Conclusion}
In this article, we proposed ScoreAD, which leverages manifold distribution discrepancy between anomalous and background spectra for hyperspectral anomaly detection. 
Specifically, ScoreAD utilizes a score-based generative model (SGM) to estimate the score of each data point in the high-dimensional space. It determines the anomaly degree of each spectrum based on the directions of the scores from its corresponding perturbed datapoints. 
The effectiveness of our proposed ScoreAD method has been demonstrated through experiments conducted on four hyperspectral datasets.

It is important to note that our method is grounded in the manifold hypothesis, which is satisfied by a wide variety of datasets. Consequently, our anomaly detection approach is theoretically applicable to other anomaly detection task that meets the corresponding conditions (where normal data lies on the manifold, but anomalous data does not). This includes anomaly detection in visible light images, industrial data, hyperspectral imagery, and so on. 
We hope that the proposed ScoreAD method will inspire researchers working on anomaly detection tasks.

\footnotesize
\bibliographystyle{IEEEtran}
\bibliography{ref}

\end{document}